\documentclass{article}
\usepackage{ijcai25}

\usepackage{times}
\usepackage{soul}
\usepackage{url}
\usepackage[hidelinks]{hyperref}
\usepackage[utf8]{inputenc}
\usepackage[small]{caption}
\usepackage{graphicx}
\usepackage{amsmath}
\usepackage{amsthm}
\usepackage{booktabs}
\usepackage{algorithm}
\usepackage{algorithmic}
\usepackage[switch]{lineno}

\usepackage{color}

\usepackage{algorithm}
\usepackage{algorithmic}
\usepackage{multirow}
\usepackage{float}
\usepackage{array}
\usepackage{amsmath}
\usepackage{amssymb}

\newcommand{\mc}[1]{\mathcal{#1}}
\newcommand{\bl}[1]{\textbf{#1}}

\newcommand{\mypara}[1]{\noindent\textbf{#1}}

\title{CD$^2$: Constrained Dataset Distillation for Few-Shot Class-Incremental Learning
}

\author{
Kexin Bao$^{1,2}$
\and
Daichi Zhang$^{1,2 \dagger}$\and
Hansong Zhang$^{1,2}$\and
Yong Li$^1$\and
Yutao Yue$^3$\And
Shiming Ge$^1$\thanks{Corresponding author; $\dag$: Project lead.}
\\
\affiliations
$^1$Institute of Information Engineering, Chinese Academy of Sciences\\
$^2$School of Cyber Security, University of Chinese Academy of Sciences\\
$^3$Hong Kong University of Science and Technology (Guangzhou)\\
\emails
\{baokexin, zhangdaichi, zhanghansong, liyong\}@iie.ac.cn,
yutaoyue@hkust-gz.edu.cn,
geshiming@iie.ac.cn
}

\begin{document}

\maketitle


\begin{abstract}
Few-shot class-incremental learning (FSCIL) receives significant attention from the public to perform classification continuously with a few training samples, which suffers from the key catastrophic forgetting problem. Existing methods usually employ an external memory to store previous knowledge and treat it with incremental classes equally, which cannot properly preserve previous essential knowledge. To solve this problem and inspired by recent distillation works on knowledge transfer, we propose a framework termed \textbf{C}onstrained \textbf{D}ataset \textbf{D}istillation~(\textbf{CD$^2$}) to facilitate FSCIL, which includes a dataset distillation module (\textbf{DDM}) and a distillation constraint module~(\textbf{DCM}). Specifically, the DDM synthesizes highly condensed samples guided by the classifier, forcing the model to learn compacted essential class-related clues from a few incremental samples. The DCM introduces a designed loss to constrain the previously learned class distribution, which can preserve distilled knowledge more sufficiently. Extensive experiments on three public datasets show the superiority of our method against other state-of-the-art competitors.
\end{abstract}

\section{Introduction}

In real-world applications such as robotics, healthcare, and remote sensing, dealing with sequential data streams and few-shot data arise frequently and often simultaneously~\cite{DBLP:journals/titb/SunZWT24,DBLP:journals/cea/LiYZLZ24}. While deep neural models demonstrate impressive performance in static and data-abundant settings~\cite{DBLP:conf/aaai/0001SYZ24,DBLP:conf/aaai/0001WZ24}, it is still significantly challenging for them to learn new concepts continually with a very restricted quantity of labeled samples. In this situation, few-shot class-incremental learning (FSCIL) has emerged as a promising solution and has attracted much research attention~\cite{tao2020cvpr}. In FSCIL, the model acquires extensive knowledge with sufficient labeled samples in the base session, and continually learns new knowledge from a few samples while retaining previously learned concepts in subsequent incremental sessions. Within FSCIL, the primary challenge lies in preventing catastrophic forgetting~\cite{DBLP:conf/iccv/KangZZWCMH23,DBLP:conf/nips/BabakniyaF0SA23}, a phenomenon where models tend to overwrite previously acquired concepts when confronted with new data due to the unavailability of data from previous sessions, especially learning with limited data.

\begin{figure}[t]
\centering
\includegraphics[width=1\linewidth]{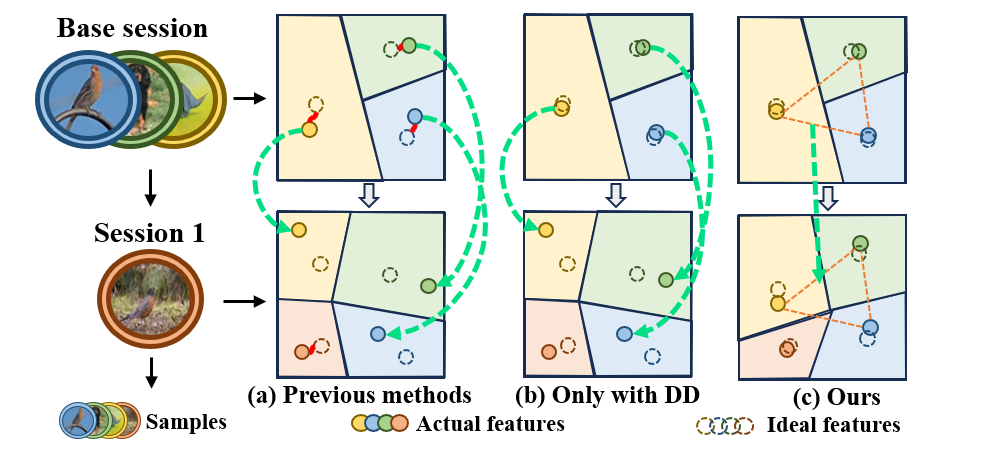}
\caption{Red lines are the gap between actual extracted features and ideal features with all critical knowledge. Green arrows are the covariate shift. (a) Previous methods build a memory that dilutes critical knowledge, and treat the memory as equal to the real data, causing the covariate shift.
(b) Using dataset distillation, the method can build a memory with more critical knowledge, but also faces the challenge of the covariate shift. (c) Our method builds and uses a memory incorporating distillation, which gains more critical knowledge and reduces the covariate shift.
}
\label{fig:motivation}
\end{figure}

Recently, mainstream methods tend to freeze the backbone and adopt an external memory as an auxiliary tool during incremental sessions to preserve previously learned knowledge~\cite{9711437,DBLP:journals/pami/YangLZLLJY23,DBLP:journals/corr/abs-2403-18550}. The memory stores a small amount of data from previous sessions by replaying real samples~\cite{9711437,9904282}, generating samples~\cite{DBLP:conf/eccv/LiuGCWYCT22,agarwal2022mm} or calculating pseudo features~\cite{DBLP:conf/cvpr/HerscheKCBSR22,DBLP:journals/tip/JiHLPL23,yang2023neural,liu2023tpami}, enabling quick recall of old knowledge in subsequent sessions. Compared to only freezing the backbone, using a memory significantly boosts performance in incremental sessions, which illustrates that the memory enhances the ability of the model to revisit old knowledge and improves its information discrimination capacity. By leveraging prior experience effectively, using the memory offers a robust solution to catastrophic forgetting in an FSCIL setting. 

However, we find that existing methods may still learn redundant features and cannot preserve the previous distribution in incremental sessions. As shown by red lines in Figure~\ref{fig:motivation} (a), when building the memory, previous methods dilute critical class-related knowledge due to mixing discriminative and redundant knowledge, which hinders the recall of old knowledge in subsequent sessions. Recently, Dataset Distillation (DD) is proposed to obtain a few highly condensed and informative samples from a dataset~\cite{DBLP:conf/iclr/ZhaoMB21,DBLP:conf/aaai/0003LW0G24}, which aligns with our goal of learning essential class-related knowledge. Therefore, we introduce DD for building a memory to obtain critical knowledge and synthesize highly informative samples as shown in Figure~\ref{fig:motivation} (b). 
Further, there is a distribution gap between data in the memory and the new dataset in incremental sessions~\cite{DBLP:conf/icml/YangZDR24}.
Treating them equally prompts the model to focus on distribution differences, leading to the covariate shift of old classes as illustrated by green arrows in Figure~\ref{fig:motivation} (a) and (b). Facing the gap, we incorporate the core idea of knowledge distillation (KD)~\cite{kd} during training. KD transfers knowledge from a teacher model (usually a well-trained, high-performing model) to a student model. And we constrain the change of distribution across continual sessions meticulously by putting the previously trained model as the teacher for the current model, which can mitigate the covariate shift and catastrophic forgetting. Combined with the core idea of distillation, our method can extract and curate critical knowledge as shown in Figure~\ref{fig:motivation} (c). 

Specifically, we propose a \textbf{C}onstrained \textbf{D}ataset \textbf{D}istillation framework \textbf{CD$^2$} to support FSCIL, which contains a more refined memory strategy and a more effective retention method. Firstly, we analyze existing popular memory strategies, including sample replay (reusing past training samples during incremental learning) and prototype computing (calculating the mean of intermediate features for each class) Then, to retain more essential knowledge, we propose a dataset distillation module  (\textbf{DDM}) inspired by DD~\cite{DBLP:conf/iclr/ZhaoMB21}. DDM synthesizes samples for each class to condense critical class-related knowledge, which retains knowledge stably and prevents performance degradation. Further, we employ a distillation constraint module (\textbf{DCM}) to transfer knowledge in a stable and flexible manner. The DCM promotes features and structures of old classes in the memory to align seamlessly between previous and current sessions, thereby enabling the model to constrain the distribution gap and utilize the memory more precisely and efficiently. 

Our main contributions can be summarized as follows: 
\begin{itemize}
    \item We propose a framework termed Constrained Dataset Distillation~(\textbf{CD$^2$}) to facilitate the FSCIL task. To the best of our knowledge, we are the first to introduce dataset distillation to FSCIL, which could effectively captures critical knowledge.
    \item We propose a dataset distillation module (\textbf{DDM}) to build a memory with more critical knowledge and a distillation constraint module (\textbf{DCM}) to reduce the covariate shift and maintain the stability of knowledge transmission, ensuring effective leveraging of dataset distillation.
    \item Extensive and comprehensive experiments on three benchmark datasets are conducted, where all results demonstrate the effectiveness of our framework.
\end{itemize}

\section{Related works}
\label{sec:formatting}


\subsection{Few-Shot Class-Incremental Learning}

Few-shot class-incremental learning (FSCIL) enables the model to incrementally learn new knowledge while effectively retaining old knowledge with a few new samples~\cite{DBLP:journals/pami/YangLZLLJY23,DBLP:journals/corr/abs-2403-18550}. Some early methods update the whole model using the topology~\cite{tao2020cvpr} and knowledge distillation~\cite{DBLP:conf/aaai/DongHTCWG21} in incremental sessions, which suffer from catastrophic forgetting and over-fitting due to finetuning a large number of parameters. 

Based on this, mainstream methods finetune part of the parameters with the assistance of a memory during incremental learning, which can reduce overfitting while retaining certain old knowledge. Some methods select random~\cite{9711437} or adaptive~\cite{9904282} samples from old classes as the memory. However, replaying a few samples cannot obtain a representative and high generalization representation, which is also prone to exposing privacy. Some generative methods effectively discourage forgetting and protect privacy by generating synthetic data with a GAN-like model~\cite{DBLP:conf/eccv/LiuGCWYCT22,agarwal2022mm}. However, there are still deviations between generated and real samples. And some methods store pseudo-features to mitigate catastrophic forgetting under privacy protection, calculating vectors from real features~\cite{DBLP:conf/cvpr/HerscheKCBSR22}, introducing virtual prototypes~\cite{zhou2022cvpr,yang2023neural,liu2023tpami}, and compressing knowledge into a small number of quantized reference vectors~\cite{chen2021iclr,DBLP:journals/tip/JiHLPL23}, which can retain knowledge quickly. However, these methods compress knowledge crudely and treat them equally with real data, affecting the model performance. Based on this, we explore a more refined memory strategy and a higher-utilization method to retain more old knowledge.

\subsection{Dataset Distillation}


Dataset Distillation (DD) aims to learning compressed data to enhance learning efficiency~\cite{DBLP:journals/corr/abs-1811-10959,DBLP:conf/cvpr/Cazenavette00EZ22a}, which is widely used in federated learning~\cite{DBLP:conf/iclr/LiuYZ23} and neural architecture search~\cite{DBLP:conf/icml/SuchRLSC20}. DD retains the critical information needed to train a model effectively, which can reduce the amount of data without significantly reducing the performance of the model.
Some optimization methods incorporate meta-learning into the surrogate image updating~\cite{DBLP:conf/nips/LooHAR22a,DBLP:conf/nips/ZhouNB22}. And other methods optimize the synthetic images by matching the training gradients~\cite{DBLP:conf/icml/LeeCJYY22}, feature distribution~\cite{DBLP:conf/cvpr/ZhaoLQY23} or training trajectories~\cite{DBLP:conf/icml/KimKOYSJ0S22,DBLP:conf/icml/LeeCJYY22}. Different from traditional data compression, DD preserves adequate task-useful information so that the model trained on it can generalize well to other unseen data.

\section{Preliminaries}
\label{sec:method}

\subsection{Formulation of FSCIL}

Few-shot class-incremental learning (FSCIL)~\cite{tao2020cvpr} consists of a base session and multiple incremental sessions, which trains a model incrementally on a time sequence of training sets $\mathcal{D}=\{\mc{D}^{(t)}\}_{t=0}^{T}$. The base ($0$-th) session dataset $\mc{D}^{(0)}$ consists of a large label space $C^{(0)}$ with sufficient samples per class. And the $t$-th incremental session dataset $\mc{D}^{(t)} (t>0)$ has limited data with its label space $C^{(t)} (t>0)$, which has $N$ classes and $K$ training examples per class (\emph{$N$-way $K$-shot}). Any two label spaces are disjoint, meaning $C^{(i)}\cap C^{(j)}=\varnothing$ for all $i, j (i \neq j)$. Among them, $\mc{D}^{(t)}=\{(\bl{x}_i,y_i)\}_{i=0}^{|\mc{D}^{(t)}|}$, where $\bl{x}_i$ is an example (e.g, image), $y_i\in C^{(t)}$ denotes its target. 
While evaluating the performance in session $t$, the model is assessed on the current and all previous validation datasets $\{\mc{T}^{(i)}\}_{i=0}^t$. 


During training, we train a model $\phi(x)$ in all sessions, which consists of a backbone $\phi_b(\bl{x}_i)$ with the intermediate feature $\bl{f}_i = \phi_b(\bl{x}_i)$, and a classifier $\phi_c(\bl{f}_i)$ with the final output $\bl{v}_i=\phi_c(\bl{f}_i)$. The classifier contains an MLP block to project intermediate features and a fully connected layer for classification. 
We train the model on dataset $\mc{D}^{(0)}$ in the base session, then finetune the classifier with dataset $\mc{D}^{(t)} (t>0)$ and an extra memory $\mathcal{M}^{(t)}$ in $t$-th incremental session. Thus, the FSCIL task is formulated into two steps as follows:
\begin{subequations}\label{eq:problem}
\begin{align}
\min&\mathbb{E}_B(\mathcal{D}^{(0)};\phi^{(0)}(\bl{x}_i)), \\
\min\mathbb{E}_I&(\mathcal{D}^{(t)}, \mathcal{M}^{(t)};\phi^{(t)}(\bl{x}_i)),
\end{align}
\end{subequations}
where Eq.~(\ref{eq:problem}a) and Eq.~(\ref{eq:problem}b) minimizes the empirical risk $\mathbb{E}_B$ in the base session ($t=0$) and $\mathbb{E}_I$ in incremental sessions ($t>0$). In incremental sessions, the model only has access to the dataset of the current session and not the training set of previous sessions, causing catastrophic forgetting. In this context, the memory plays an important role in mitigating this issue. Previous methods always build and use memory crudely, which cannot properly preserve critical old knowledge and further affects the distinguishing ability of the model. Therefore, we are devoted to exploring a more refined memory strategy and a higher-utilization method, which preserves more old knowledge during incremental learning and improves the continuous learning ability of the model.

\begin{figure}[t]
\centering
\includegraphics[width=0.8\linewidth]{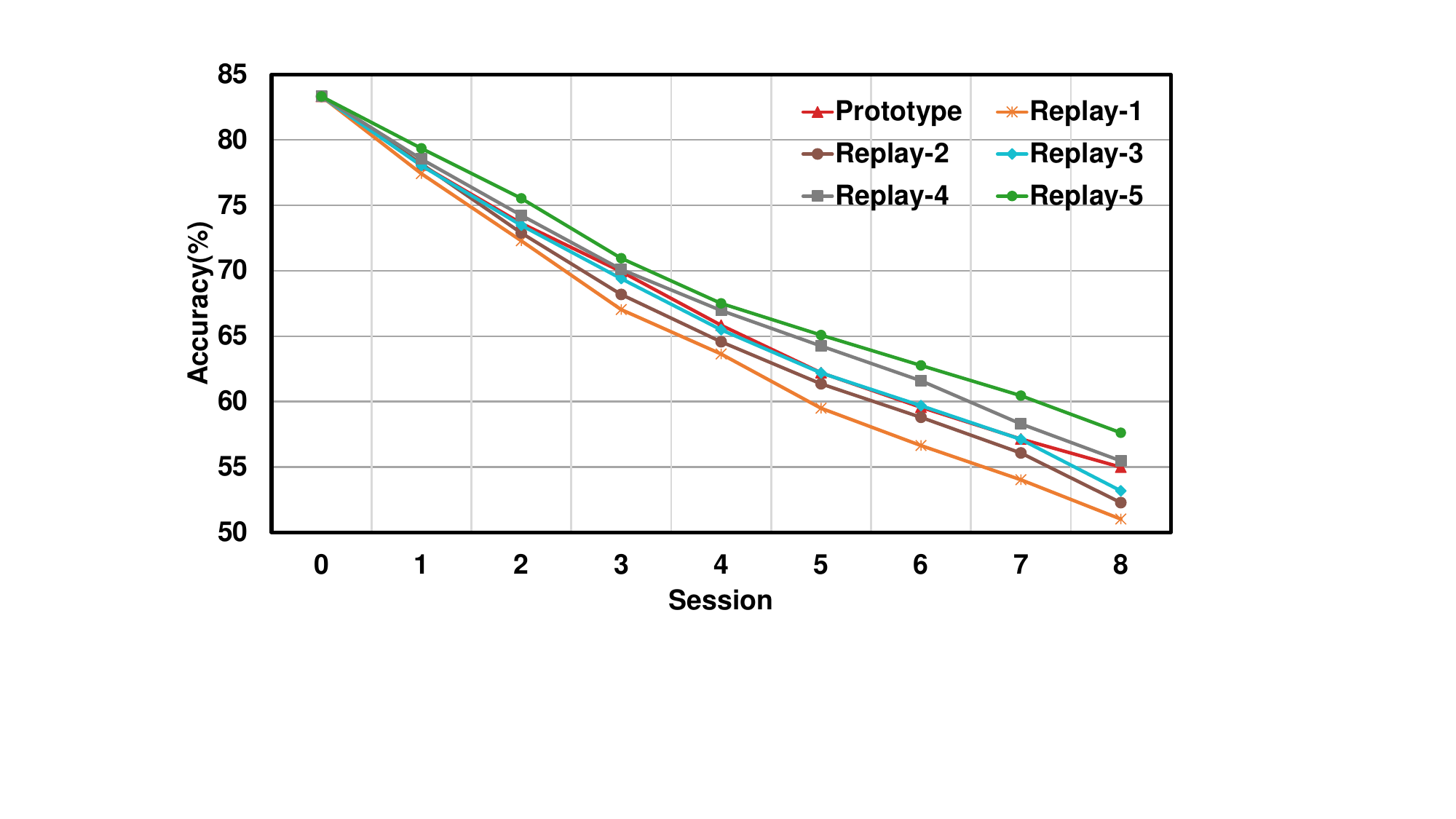}
\caption{The downtrend of accuracy for all sessions on CIFAR100 as an example. "Replay-1``, "Replay-2``, "Replay-3``, "Replay-4``, and "Replay-5`` denote using sample replay to build a memory with 1, 2, 3, 4, and 5 images per class when training a model, which trains the model by optimizing Eq.~\ref{eq:replay}. And "Prototype`` means using prototypical features as a memory as Eq.~\ref{eq:prototype} when training a model. }
\label{fig:analyse}
\end{figure}


\subsection{Explore the Previous Memory Strategies}

\mypara{Sample replay.} To achieve high performance in the joint space of old and new classes, the sample replay technique is used in FSCIL. Some methods store some real samples from previous sessions in a memory. In $t$-th session ($t>0$), few samples from dataset $\mc{D}^{(t-1)}$ are sent to the memory $\mathcal{M}^{(t)}$, which alongside new added samples to finetune the model as

\begin{equation}\label{eq:replay}
\begin{aligned}
~\mc{L}=~\mc{L}_{ce}(\phi(\bl{x}_i),y_i), ~(\bl{x}_i,y_i) \in \{\mc{D}^{(t)} \cup \mathcal{M}^{(t)}\}.
\end{aligned}
\end{equation}

These methods re-expose previous samples to the model during training, which is like a reminder to help the model retain previous knowledge when it learns new knowledge.

\mypara{Prototype computing.} Although using sample replay can recall old knowledge, it may develop biases for the model due to limited knowledge and is prone to exposing privacy, which restricts the ability to mitigate catastrophic forgetting. Inspired by ProtoNet~\cite{snell2017nips}, some methods store prototypes as a memory for subsequent tasks as:
\begin{equation}
    \label{eq:prototype}
    \begin{aligned}
    \overline{\bl{f}}_{c}=\frac{1}{|\mc{D}(y_i=c)|}\sum\nolimits_{\bl{x}_i \in \mc{D}(y_i=c)}\phi_b(\bl{x}_i), c \in  C^{(t)}.
    \end{aligned}
\end{equation}


Each prototype is the mean of intermediate features, condensing the common pattern of each class within an embedding vector. The prototypes remain constant across incremental sessions to alleviate catastrophic forgetting.


\mypara{Analyse.} Figure~\ref{fig:analyse} shows the performance of building a memory by sample replay and prototype computing. When using sample replay, the performance gradually increases with the number of samples, which indicates that more samples cover more knowledge. However, the selection problem of samples, resource consumption, and privacy issues cannot be ignored. When outlier samples are chosen for replay, it affects the model to generalize and retain old knowledge properly. Prototype computing preserves privacy and integrates knowledge efficiently. However, averaging features may cause the dilution of critical class-related information, which fails to capture the knowledge best suited for model training. Besides, they may be influenced by a few outlier samples with extreme values, especially when facing scarce data in incremental sessions. These outliers introduce biases and noise that can compromise the model performance. To address that, we further introduce a more available strategy to build the memory and a more efficient method to use the memory.

\section{Methodology}

\begin{figure*}[t]
\centering
\includegraphics[width=1\linewidth]{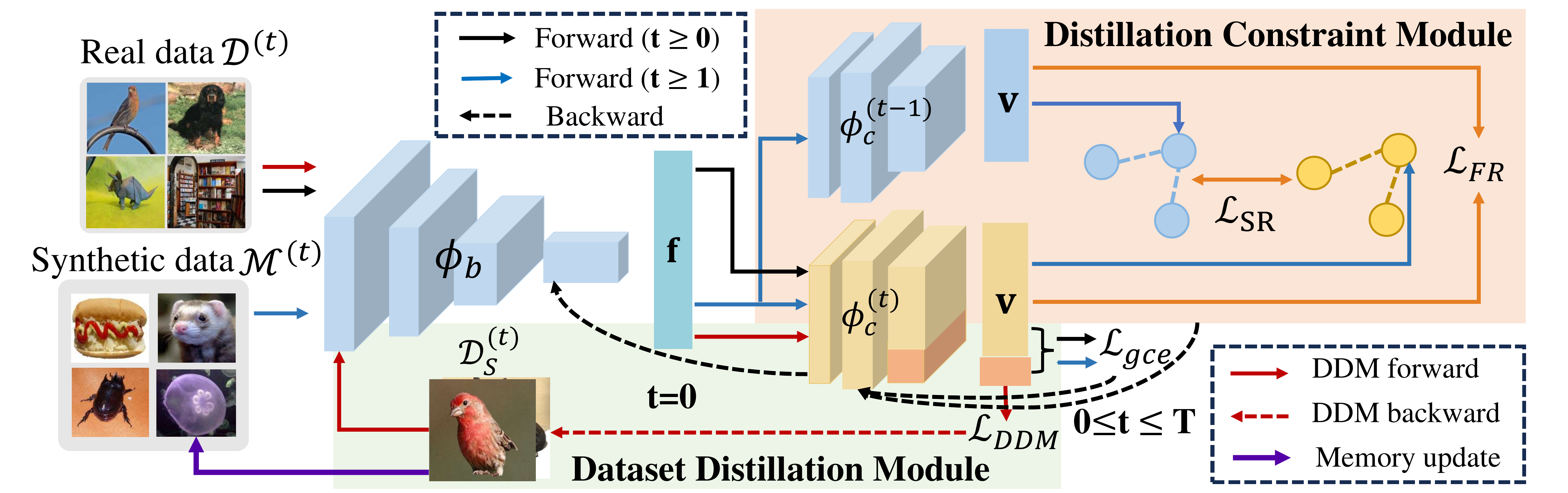}
\caption{The framework \textbf{CD$^2$} generates memory by a \textbf{dataset distillation module} (DDM) and finetunes the model with a \textbf{distillation constraint module} (DCM), containing a backbone $\phi_b$, a classifier $\phi_c$, and an extra memory $\mathcal{M}^{(t)}$. During model training, we train the model on $\mc{D}^{(0)}$ in the base session ($t=0$), then freeze the backbone and finetune the classifier with DCM on dataset $\mc{D}^{(t)}$ and the memory $\mathcal{M}^{(t)}$ in $t$-th incremental session ($t>0$). And after model training for each session ($t<T$), we synthetic samples as $\mc{D}^{(t)}_{S}$ by DDM and send them to the memory $\mathcal{M}^{(t+1)}$.
}
\label{fig:method}
\end{figure*}

As shown in Figure~\ref{fig:method}, our CD$^2$ framework for FSCIL begins with training a model with sufficient data in the base session. In incremental sessions, we finetune the classifier with the assistance of a memory. Before each incremental session, we generate a set by a dataset distillation module (DDM) as close as possible to critical knowledge and store it in the memory, which can be trained together with the current dataset to achieve the purpose of learning new concepts and preserving old concepts. In addition to the cross-entropy loss, we also employ a distillation constraint module (DCM) to mitigate catastrophic forgetting while balancing stability and flexibility to make better use of the generated data. 

\subsection{Dataset Distillation Module}

Although previous memory strategies mitigate catastrophic forgetting effectively, they compress knowledge crudely. These strategies lump together critical knowledge that enhances the discriminative ability of the model and redundant knowledge that contributes less to the model performance. Due to a lack of guidance in obtaining critical knowledge, these methods inadvertently downplay critical class-related knowledge and even overlook essential details. The dilution of critical knowledge not only impairs the model performance and discrimination ability, but also limits its potential for continuous learning and self-optimization from experience. 

Based on this, we are oriented to effectively retain more critical old knowledge and propose a dataset distillation module (DDM). The core of the strategy is \textit{\textbf{to choose critical class-related knowledge that is valid for incremental sessions}}, enhancing the adaptability and recognition ability of the model under FSCIL settings. In this way, the model can keep and update knowledge that is critical to performance, while effectively weakening less important knowledge.

Inspired by dataset distillation~\cite{DBLP:conf/iclr/ZhaoMB21,DBLP:conf/iccv/LiuW23}, we synthesize samples for each class to compress critical knowledge. We randomly select $K$ samples from $\mc{D}^{(t)}$ for each class and initialize the synthetic set $\mc{D}_S^{(t)} = \{(\bl{x}_m,y_m)|y_m \in C^{(t)} \}_{m=0}^{|C^{(t)}| \times K}$. While synthesizing, we freeze the whole model and update the synthetic set by class. The maximum mean discrepancy ($\bl{MMD}(\cdot)$)~\cite{DBLP:journals/jmlr/GrettonBRSS12} is used to estimate the distance between the real and synthetic data distribution as:

\begin{equation}\label{eq:simulation}
\begin{split}
&\bl{MMD}(\phi_b,\mc{D}_S,\mc{D}^{(t)}) = \sup (E(\mc{D}_S)-E(\mc{D}^{(t)}))\\
& = (\frac{1}{|\mc{D}_S^{(t)}|}\sum\phi_b(\bl{x}_m)-\frac{1}{|\mc{D}^{(t)}|}\sum\phi_b(\bl{x}_i))^2,
\end{split}
\end{equation}
where $\sup$ is taking an upper bound and $E$ is the expectation.


To synthesize samples more finely, we convert Eq.~(\ref{eq:simulation}) to estimate the distance of synthetic set $\mc{D}_S^{(t)}$ and real dataset $\mc{D}^{(t)}$ and optimize the loss for each class as follows
\begin{equation}\label{eq:s2}
\begin{split}
&~\mc{L}_{DDM} = \sup (\mc{D}^{(t)}_S(y_m = c)-\mc{D}^{(t)}(y_i = c))\\
& = (\frac{1}{K}\sum_{y_m = c}\phi(\bl{x}_m)-\frac{1}{|\mc{D}^{(t)}(y_i = c)|}\sum_{y_i=c}\phi(\bl{x}_i))^2,
\end{split}
\end{equation}
where $c \in C^{(t)}$. The learned synthetic set can gain complementary information and summarize important knowledge from different samples of the same class, which helps improve the power of statistical analyses and ignores sampling errors. Meanwhile, it reduces the possibility of privacy leakage because it is difficult to recover the source data. Further, to reduce training consumption while gaining critical knowledge, we replace the dataset $\mc{D}^{(t)}$ with dataset $\mc{D}_R^{(t)}$ sampled from dataset $\mc{D}^{(t)}$ to perform the synthesis and calculation.
\subsection{Distillation Constraint Module} 


Facing catastrophic forgetting, we involve a memory generated by DDM during training to recall old knowledge. Generally, a synthetic sample in the memory has a different distribution and contains more abundant knowledge than a real sample. The same treatment brings the covariate shift of old classes when adding new classes, and the classification loss cannot take full advantage of rich knowledge, impacting the model performance and the discrimination ability.

Based on this, we incorporate a distillation constraint module (DCM) in incremental sessions \textit{\textbf{to limit the distribution change across continual sessions, thereby bridging the distribution gap between data in the memory and the dataset}}, which facilitates the discriminative ability between old and new classes. 
The DCM consists of feature retention loss and structure retention loss, which ensure that old knowledge is as accurately retained as possible.

\textbf{Feature retention loss (FR loss):} The feature retention (FR) loss guarantees the consistency of features for old classes between the previous and the current session, thereby maintaining the integrity and coherence of processing data across different sessions. While calculating the FR loss, we get pairs of vectors $\{\bl{v}^{(t-1)}_m, \bl{v}^{(t)}_m\}_{m=0}^{|\mathcal{M}^{(t)}|}$ from the memory $\mathcal{M}^{(t)}$, where $\bl{v}^{(t-1)}_m = \phi_c^{(t-1)}(\bl{f}_m)$ and $\bl{v}^{(t)}_m = \phi_c^{(t)}(\bl{f}_m)$. And we constrain the first $C = |\sum_{i=0}^{t-1}C^{(i)}|$ elements of each pair of vectors to be consistent, which is formulated as 
\begin{equation}
\mathcal{L}_{FR}= \frac{1}{\mathcal{M}^{(t)}}\sum_{(x_m,y_m)\in \mathcal{M}^{(t)}} \left|\bl{v}^{(t-1)}_m-\bl{v}^{(t)}_m[:C]\right|.
\label{eq:fr}
\end{equation}

With the FR loss, the pairs of vectors in the previous model $\phi_c^{(t-1)}$ and the current model $\phi_c^{(t)}$ exhibit a tendency to convergence. And the FR loss constrains the location information, which enables the gap between real data and synthetic data as constant as possible to maintain the coherence and stability of knowledge transmission. 

\textbf{Structure retention loss (SR loss):} Further, we hope to retain old knowledge more flexibly and elastically. Inspired by relational knowledge distillation (RKD)~\cite{DBLP:conf/cvpr/ParkKLC19}, we employ the structure retention (SR) loss to ensure the consistency of the structure distribution of synthetic samples in the previous session and the current session. RKD transfers structural information among a set of samples from the teacher model to the student model, endowing the student model with more flexibility to learn new knowledge. It constrains structural information by constraining the angle between triplets of samples $\{(x_a,x_b,x_c)\}$, which is as follows
\begin{equation}
\mathcal{L}_{RKD}= \sum_{\{(x_a ,x_b ,x_c )\}} |\cos \angle t_a t_b t_c - \cos \angle s_a s_b s_c|,  
\label{eq:rkd}
\end{equation}
where $t_i = \psi(x_i)$ and $s_i = \zeta(x_i)$. Here, $\psi(x_i)$ is the teacher model and $\zeta(x_i)$ is the student model.


We incorporate the core idea of RKD into our CD$^2$, transferring the structural information of the memory (synthetic samples) in the feature space from the old model $\phi_c^{(t-1)}$ to the current model $\phi_c^{(t)}$. To constrain the feature structure, we formulate the SR loss based on a linear transformation between the features, which can be expressed as:
\begin{equation}
\mathcal{L}_{SR}= \frac{1}{\mathcal{M}^{(t)}}\sum_{(x_m,y_m)\in \mathcal{M}^{(t)}} \left|\bl{P}^{(t-1)}-\bl{P}^{(t)}[:C]\right|,
\label{eq:sp}
\end{equation}
where $\bl{P}^{(t)} = \bl{v}_m \bl{V}^{T}$ is a vector of linear transformation and $\bl{V} = \{\phi_c^{(t)}(\bl{f}_m)|(x_m,y_m)\in \mathcal{M}^{(t)}\}$ is a matrix of output vectors in the memory $\mathcal{M}^{(t)}$.
By minimizing this loss, we not only ensure the stability of the model when dealing with previous knowledge but also maintain its capacity for exploration and adaptation to new data and knowledge. 


\mypara{Distillation constraint module (DCM).} The above two components form the distillation constraint module (DCM), maintaining a stable identification of old classes while being more flexible. The total loss can be summarized as follows:
\begin{equation}
\mathcal{L}_{DCM}= \alpha \mathcal{L}_{SR} + \beta \mathcal{L}_{FR},
\label{eq:rp}
\end{equation}
where $\alpha$ and $\beta$ are factors. The FR loss functions to constrain location information, while the SR loss aims to preserve structural information. And these dual constraints help to stabilize the distribution changes across continuous learning sessions when incorporating new classes. The DCM is designed to subtly account for the two core requirements of stability and flexibility, which reduces the covariate shift. 

\subsection{Model Training \& Memory}

Fig.~\ref{fig:method} illustrates the training process of our CD$^2$ in $t$-th session. According to different inputs, we divide the whole training into two stages: base learning and incremental learning. 

\textbf{Base learning:} \textit{During model training,} we train a model with dataset $\mc{D}^{(0)}$ with a sufficient amount of samples. For each training example $\bl{x}_i$, the model $\phi(\bl{x}_i)$ represents it into a final vector $\bl{v}_i$. In the base session, both the backbone $\phi_b(\bl{x}_i)$ and the classifier $\phi^{(0)}_c(\bl{f}_i)$ are jointly trained on $\mc{D}^{(0)}$ by minimizing the empirical risk loss as
\begin{equation}
    \mathcal{L}_{b} = \frac{1}{|\mc{D}^{(0)}|}\sum_{(\bl{x}_i,y_i) \in \mc{D}^{(0)}} \mathcal{L}_{ce}(\phi^{(0)}(\bl{x}_i),y_i).
\label{eq:baseloss}
\end{equation}

\textit{After model training}, we build a small buffer as a memory to participate in subsequent training. We select the initial samples from dataset $\mc{D}^{(0)}$ for each class and generate samples as Eq.~(\ref{eq:s2}) to gain a set $\mc{D}_S^{(0)}$. And we send the set $\mc{D}_S^{(0)}$ to the memory $\mathcal{M}^{(1)}$.

\textbf{Incremental learning:} \textit{During model training,} we freeze the backbone and finetune the classifier with dataset $\mc{D}^{(t)}$ and the memory $\mathcal{M}^{(t)}$. In $t$-th incremental session, we minimize the learning objective function as:
\begin{equation}
\begin{split}
\mathcal{L}_{i} &= \mathcal{L}_{gce} 
 + \mathcal{L}_{DCM}\\
 &= \mathcal{L}_{gce} 
 +\alpha \mathcal{L}_{SR} + \beta \mathcal{L}_{FR},
\end{split}
\label{eq:incloss}
\end{equation}
where $\mathcal{L}_{gce} $ is the global classification loss as
\begin{equation}
\begin{split}
 \mathcal{L}_{gce} &= \frac{1}{|\mc{D}^{(t)}|}\sum_{(\bl{x}_i,y_i) \in \mc{D}^{(t)}} \mathcal{L}_{ce}(\phi^{(t)}(\bl{x}_i),y_i)\\
+&\frac{1}{|\mathcal{M}^{(t)}|}\sum_{(\bl{x}_m,y_m) \in \mathcal{M}^{(t)}} \mathcal{L}_{ce}(\phi^{(t)}(\bl{x}_m),y_m).
\end{split}
\label{eq:gce}
\end{equation}


As the number of sessions increases, the ratio of previous classes to new classes gets larger, gradually making it more difficult to retain previous knowledge. The model should allocate more energy to the old knowledge as the session grows. Meanwhile, considering that the excessive proportion of the FR loss will affect the flexibility of the model, we only constrain the SR loss adaptively, which factors are set as $\alpha = \ln((-50/{|\sum_{i=0}^{(t)}C^{(i)}|})^3+2)$.

\textit{After model training}, we generate a synthetic set $\mc{D}_S^{(t)}$ from dataset $\mc{D}^{(t)}$ in a similar way as in base learning and add it to the memory $\mathcal{M}^{(t+1)}$ for subsequent fine-tuning.





\begin{table*}[t]
\centering
\resizebox{2.1\columnwidth}{!}{
\begin{tabular}{cccccccccccc} 
\hline
	\multirow{2}{*}{Method} & \multicolumn{9}{c}{Accuracy in each session($\%$)$\uparrow$} & Average & Average \\
    &  0 & 1 & 2 & 3 & 4 & 5 & 6 & 7 & 8 & accuracy & improvement\\
	\hline
	iCaRl \cite{rebuffi2017cvpr} & 64.10 & 53.28 & 41.69 & 34.13 & 27.93 & 25.06 & 20.41 & 15.48 & 13.73 & 32.87 & +35.69    \\
        EEIL \cite{DBLP:conf/eccv/CastroMGSA18}&	64.10&53.11 &43.71 	&35.15 &28.96& 	24.98 	&21.01 &	17.26 &	15.85 &	33.79 &	+34.77   \\
        \hline
     
        SoftNet \cite{DBLP:conf/iclr/KangYMHY23} &72.62 &67.31 &63.05 &59.39 &56.00 &53.23 &51.06 &48.83 &46.63 &57.57 &+11.10 \\
        MCNet \cite{DBLP:journals/tip/JiHLPL23} &  73.30 & 69.34 & 65.72  &61.70 & 58.75 & 56.44 & 54.59 & 53.01 & 50.72 & 60.40 & +8.27  \\  
        GKEAL \cite{zhuang2023cvpr} &74.01 &70.45 &67.01 &63.08 &60.01 &57.30 &55.50 &53.39 &51.40 &61.35 & +7.32 \\
        FACT\cite{DBLP:conf/cvpr/0001WYMPZ22} &74.60&72.09 &67.56 &63.52&61.38 &58.36 &56.28&54.24 &52.10 &62.24&+6.43 \\
        C-FSCIL \cite{DBLP:conf/cvpr/HerscheKCBSR22} & 77.47 & 72.40 & 67.47 & 63.25 & 59.84 & 56.95 & 54.42 & 52.47 & 50.47 & 61.64 & +7.03  \\ 
        MICS~\cite{DBLP:conf/wacv/KimJPY24}& 78.18& 73.49 &68.97 &65.01& 62.25 &59.34 &57.31 &55.11 &52.94&63.62&+5.05\\
        ALICE \cite{DBLP:conf/eccv/PengZWLL22} & 79.00 & 70.50 & 67.10 & 63.40 & 61.20 & 59.20 & 58.10 & 56.30 & 54.10 & 63.21 & +5.46\\
        CABD \cite{10203568}&79.45&75.38 &71.84 &67.95 &64.96 &61.95 &60.16 &57.67 &55.88 &66.14 &+2.53 \\
        OrCo~\cite{DBLP:journals/corr/abs-2403-18550} & 80.08 &  68.16 & 66.99 &60.97 &59.78 &58.60 &57.04 &55.13 &52.19 &62.11& +6.56\\
        WaRP \cite{kim2023iclr} & 80.31 &75.86 &71.87 &67.58 &64.39 &61.34 &59.15 &57.10 &54.74 &65.82 &+2.85  \\
        NC-FSCIL \cite{yang2023neural} & 82.52 & 76.82 & 73.34 & 69.68 & 66.19 & 62.85 & 60.96 & 59.02 & 56.11 & 67.50 & +1.17\\
        Revisting-FSCIL~\cite{tang2024eccv} & 82.90 & 76.30 &72.90 &67.80 &65.20 &62.00 &60.70 &58.80 &56.60 &67.02 &+1.65\\
        \hline
	 \textbf{ CD$^2$ (Ours)} & 
\textbf{83.32} & \textbf{79.42} & \textbf{74.96} & \textbf{70.33} & \textbf{67.28} & \textbf{64.21} & \textbf{61.35} & \textbf{59.81} & \textbf{57.36} & \textbf{68.67} & - \\
\hline
\end{tabular}
}
\caption{FSCIL performance on CIFAR100. ``Average accuracy'' means the average accuracy of all sessions and ``Average improvement'' calculates the improvement of our approach over other approaches. These approaches include class-incremental learning with FSCIL setting and FSCIL methods. The best results are in bold. }
\label{table1}
\end{table*}

\begin{figure}[ht]
\centering
\includegraphics[width=1.0\linewidth]{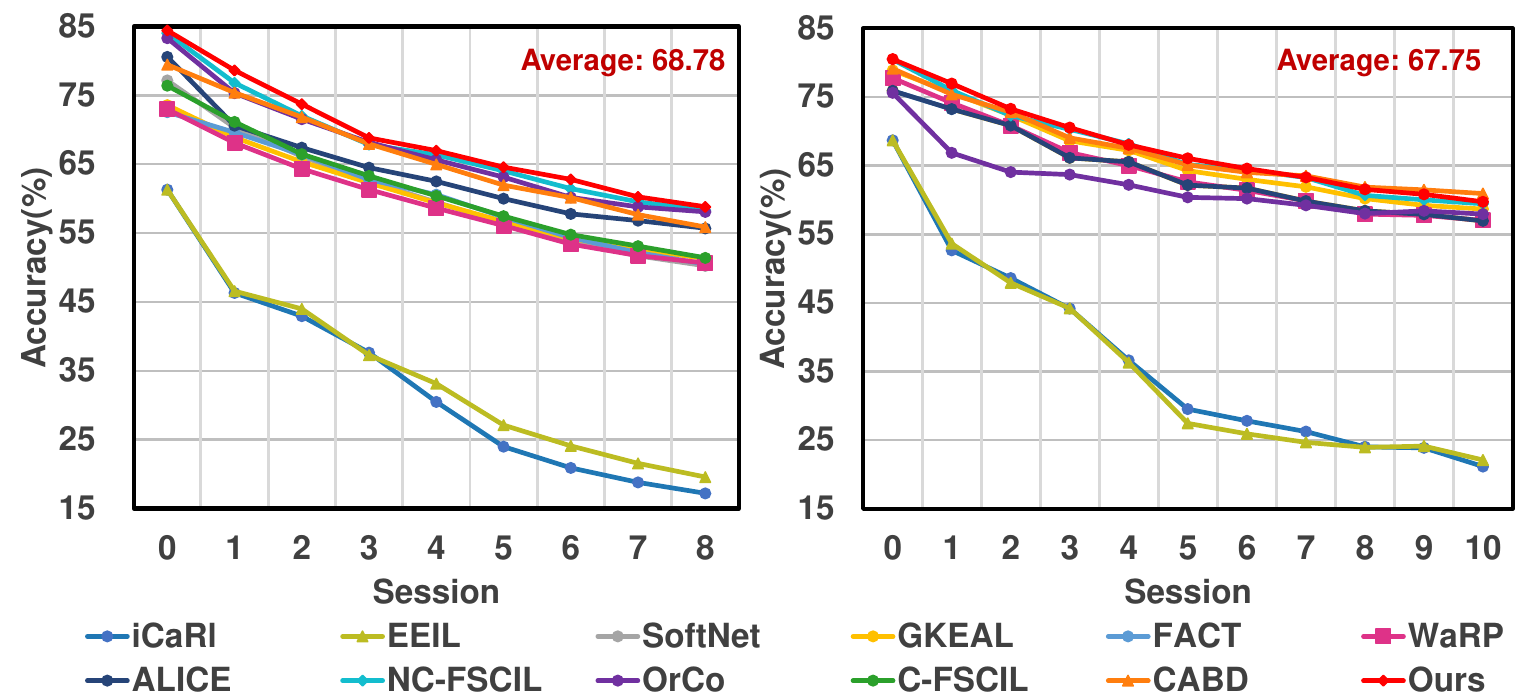}
\caption{Performance curves of our method compared to recent SOTA methods on mini-ImageNet and CUB200. Left: mini-ImageNet. Right: CUB200. ``Average'' denotes the average accuracy of all sessions. Please refer to the appendix for more details.}
\label{fig:result}
\end{figure}

\section{Experiments}
\label{sec:exper}

\subsection{Experimental Setup}

\mypara{Datasets.} We evaluate our CD$^2$ on three FSCIL benchmark datasets: CIFAR100~\cite{2009Learning}, mini-ImageNet~\cite{russakovsky2015ijcv}, and CUB200~\cite{Wah2011TheCB} following previous settings~\cite{DBLP:conf/eccv/LiuGCWYCT22,yang2023neural}. On CIFAR100 and mini-ImageNet, the 100 classes are organized into 60 base classes and 40 incremental classes. 60 base classes contain 500 training images for the training model. And 40 incremental classes are further structured in 8 different sets with a \emph{$5$-way $5$-shot} setting. And 200 classes of CUB200 are organized into 100 base classes and 100 incremental classes in a \emph{$10$-way $5$-shot} FSCIL setting for 10 incremental sessions. 

\mypara{Network.} Following previous work~\cite{DBLP:conf/eccv/LiuGCWYCT22,yang2023neural,DBLP:journals/corr/abs-2403-18550}, we employ ResNet~\cite{he2016cvpr} as a backbone. We use ResNet12 without pretraining for CIFAR100 and MiniImageNet, and ResNet18 pre-trained on ImageNet for CUB200. We adopt a three-layer MLP block and a fully connected layer as a classifier. 

\mypara{Implementation details.} Our model is optimized using SGD with momentum and adopts a cosine annealing strategy for the learning rate during training. In the base session, we train for $100$ to $200$ epochs while initializing a learning rate of $0.25$ for CIFAR100 and mini-ImageNet, and $0.01$ for CUB200. In each incremental session, we train for $100$ to $300$ iterations initializing a learning rate of $0.25$ for CIFAR100 and MiniImageNet, and $0.001$ for CUB200. Augmentations include random resizing, random flipping, Mixup~\cite{zhang2018iclr}, and CutMix~\cite{yun2019iccv}. And we set $\beta = 0.1$ for model training. When using DDM, we train $1000$ iterations initializing a learning rate of $0.2$. 


\subsection{State-of-the-art Comparison}

We evaluate our CD$^2$ on three public datasets and conduct a comparative analysis with some class-incremental learning methods with FSCIL setting and some state-of-the-art FSCIL methods using different memory strategies. Table~\ref{table1} presents the results obtained on the CIFAR100, while Figure~\ref{fig:result} illustrates the evaluation results on the mini-ImageNet and CUB200. The results demonstrate our method has superior performance across all three datasets, surpassing previous state-of-the-art methods.


Notably, our method achieves the highest average accuracy in three datasets. Firstly, our method outperforms class-incremental learning (CIL) methods(such as iCaRL~\cite{rebuffi2017cvpr}, EEIL~\cite{DBLP:conf/eccv/CastroMGSA18}), which primarily focus on mitigating catastrophic forgetting in the context of CIL rather than FSCIL. In contrast to these methods, our approach is tailored specifically to tackle the challenges of catastrophic forgetting in FSCIL. And secondly, compared with some FSCIL methods, we obtain the best accuracy in all sessions on CIFAR100 and MiniImageNet and the best average accuracy on three datasets, maintaining the accuracy advantage. Our accuracy curve declines more slowly than other methods on CIFAR100 and MiniImageNet, particularly achieving improvements of $0.80\%$ in the base session and $2.60-0.39\%$ in incremental sessions than NC-FSCIL~\cite{yang2023neural} on CIFAR100. Although the accuracy curve appears dented on several sessions on CUB200, we still achieve the highest average accuracy. The excellent performance indicates the efficacy of our CD$^2$, and the memory facilitates the preservation of previous knowledge for subsequent tasks effectively.

\subsection{Ablation Study}

To validate the effectiveness of each component, we further study the effect of different memory strategies and the DCM based on CIFAR100. More studies are in the appendix.

\mypara{The effect of different memory strategies.} We compare the impact of different memory strategies on accuracy as shown in Table~\ref{ab}. Compared with prototype computing, CD$^2$ achieves higher performance in incremental sessions, which improves more than $0.44\%$ in the first incremental session and more than $1.33\%$ in the last incremental session. Different memory strategies give different knowledge to the model, which directly affects the update direction of the parameters and the final accuracy. DDM can capture the key knowledge and is different from real data, which gives the model a good foundation to learn knowledge continually. 

\begin{table}[t]
\centering
\begin{tabular}{ccccccc}
\hline
\multirow{2}{*}{$\mathcal{L}_{SR}$} & \multirow{2}{*}{$\mathcal{L}_{FR}$}& \multirow{2}{*}{Base}& \multicolumn{2}{c}{~Prototype~} & \multicolumn{2}{c}{~DCM~}\\
     &  & &First & Final  & First & Final \\
	\hline
 & &	\multirow{4}{*}{83.32} &78.07 & 55.28 &79.01 &56.31\\
$\checkmark$ & & &78.33 & 55.72	&79.24 &56.83\\
 &	$\checkmark$& &78.57 & 55.45 &79.37 &56.06\\
\textbf{$\checkmark$} &	\textbf{$\checkmark$}& & \textbf{78.98} & \textbf{56.03}&\textbf{79.42} &\textbf{57.36}\\
\hline
\end{tabular}
\caption{The effect of different memory strategies and the DCM on CIFAR100. ``Base'' denotes the accuracy of the base session. ``First'' and ``Final'' refer to the accuracy of the first and last incremental session, respectively. The best results are in bold.}
\label{ab}
\end{table}

\mypara{The effect of the DCM.} The DCM plays an important role in the whole training. As shown in Table~\ref{ab}, the model achieves higher accuracy when both SR loss and FR loss are applied. When using the SR loss alone, the model retains old knowledge stably, but may lead to displacement. When only using the FR loss, the model can retain old knowledge more accurately, but strict requirements limit the adaptation of the model to new classes. When SR and FR losses are used together, the model can subtly pursue both flexibility and accuracy, resulting in optimal overall performance.

\subsection{Further Discussion}

For a detailed analysis, we observe the number of synthetic sets and the visualization. More studies are in the appendix. 

\mypara{The effect of $K$.} As shown in Table~\ref{num}, the model gets an improvement in all sessions as $K$ increases, which means the number of generated samples affects the accuracy of the model. As the sample size increases, the model can be exposed to more diverse data and situations, thus accumulating richer knowledge, which is directly reflected in the improved accuracy in all sessions. However, when $K \geqslant 2$, the number of samples has a negligible impact on the model performance. Considering the resource consumption and the performance, we set $K=2$.

\begin{table}[t]
\centering
\begin{tabular}{ccccc}
\hline
$K$ & Base & First & Final  & Average \\
	\hline
1 & \multirow{5}{*}{83.32} & 78.85 & 56.67 &67.74\\
2 & & 79.42 & 57.36 & 68.67\\
3 & & 79.36 &57.13 & 68.59\\
4 & & 79.45 & 56.92 & 68.23\\
\textbf{5} & &\textbf{79.48} &\textbf{57.41} &\textbf{68.70}\\
\hline
\end{tabular}
\caption{The effect of the number in synthetic samples, which is controlled by $K$. ``Base'' denotes the accuracy of the base session. ``First'' and ``Final'' refer to the accuracy of the first and last incremental session. ``Average'' is the average accuracy of all sessions. The best results are in bold.}
\label{num}
\end{table}

\begin{figure}[t]
\centering
\includegraphics[width=0.84\linewidth]{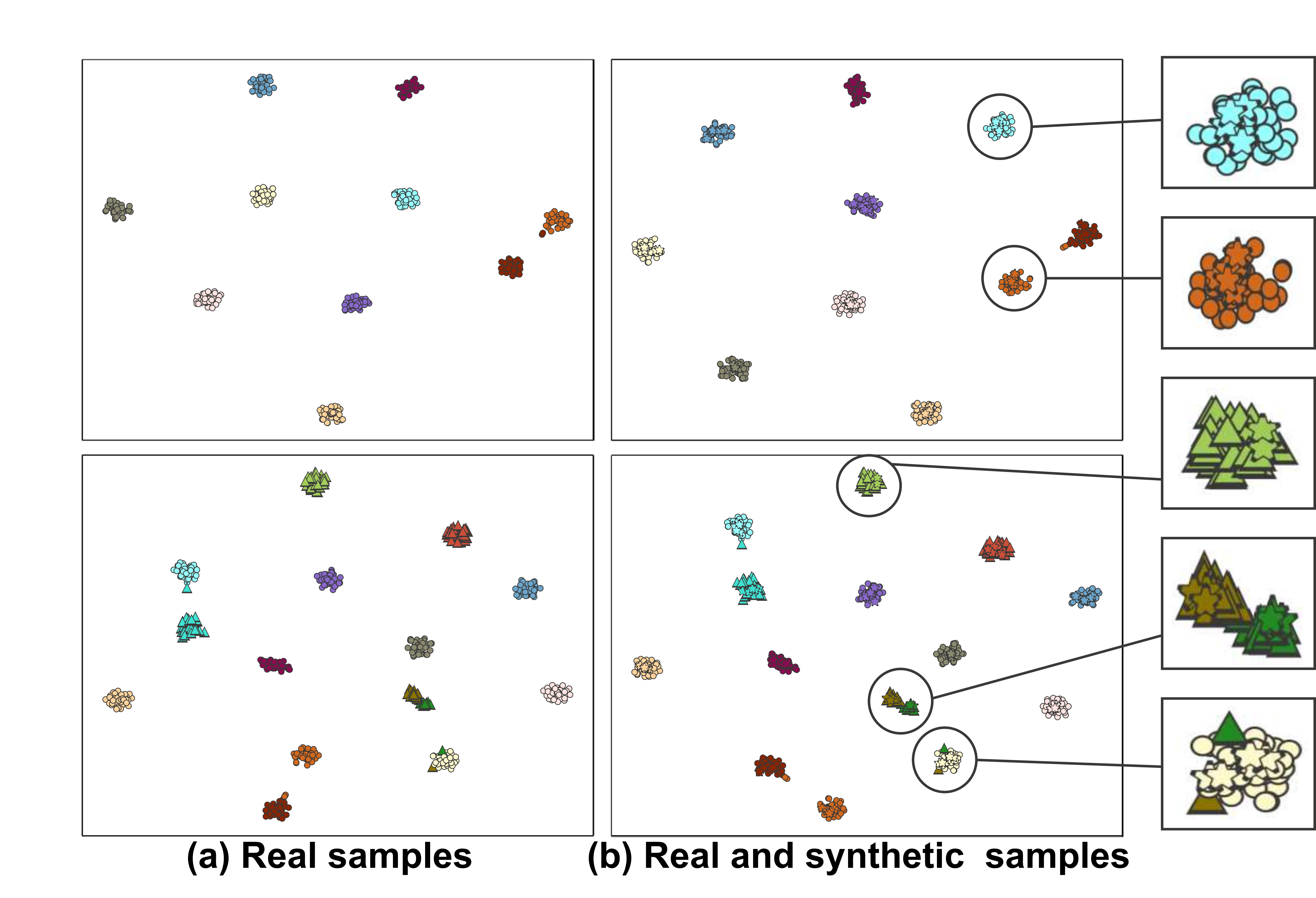}
\caption{The t-SNE visualization of representations, which uses the base session and an incremental session on CIFAR-100 as an example. We randomly select 50 examples over 10 base classes and 5 incremental classes to show the effect. And we visualize all synthetic samples from all selected classes. Symbols like `$\bullet$' and `$\blacktriangle$' represent samples of base classes and incremental classes. `$\bigstar$'  represents synthetic samples. (a) is visual features of real samples, and (b) is visual features of both real and synthetic samples. The top row shows the visualization of the base session, and the bottom row shows the visualization of the incremental session.}
\label{fig:tsne}
\end{figure}

\mypara{Representation visualization.} Fig.~\ref{fig:tsne} clearly shows the aggregation of classes by t-SNE visualization~\cite{2008Visualizing}. Firstly, the samples of the same class can be aggregated well except for some special cases, where the model can effectively capture common patterns of the same class and distinguish the intrinsic properties of the different classes. Secondly, novel classes have closer inter-class distances than base classes, which means learning new data is more difficult than learning base data, resulting in the model being less sensitive to the differences between novel classes. Finally, synthetic samples are well aggregated in their classes, where the model is capable of learning from real data and effectively transferring this learning to the synthetic set. Meanwhile, this also indicates that the model successfully captures the key characteristics of the real data and saves them in simulated samples for subsequent tasks. 

\section{Conclusion}
\label{sec:con}
In this paper, we propose a framework CD$^2$ to support FSCIL, which contains a dataset distillation module (DDM)  and a distillation constraint module (DCM). Inspired by dataset distillation, DDM synthesizes highly condensed samples with class-related clues, which can capture essential information. Furthermore, we introduce DCM to constrain distribution in incremental sessions, enabling more precise memory utilization by constraining feature alignment and structure preservation flexibly and stably, which mitigates catastrophic forgetting. 
Extensive experiments show that our approach achieves and even surpasses state-of-the-art performances.

\subsubsection{Acknowledgements}
This work was partially supported by grants from the Pioneer R$\&$D Program of Zhejiang Province (2024C01024), and Open Research Project of National Key Laboratory
of Science and Technology on Space-Born Intelligent Information Processing (TJ-02-
22-01).

\bibliographystyle{named}
\bibliography{ijcai25}

\end{document}